\documentclass[11pt]{article}
\usepackage{coling2020}
\usepackage{times}
\usepackage{url}
\usepackage{latexsym}
\usepackage{multirow}
\usepackage{amsmath}
\usepackage{graphicx}
\usepackage{placeins}
\usepackage{wrapfig}
\usepackage{bm}
\usepackage{booktabs}
\usepackage{makecell}
\usepackage{xcolor}
\usepackage{subfig}

\colingfinalcopy 

\title{ScopeIt: Scoping Task Relevant Sentences in Documents}

\author{Vishwas Suryanarayanan\thanks{{ Equal Contribution}} \\\And
  Barun Patra\footnotemark[1] \\\And Pamela Bhattacharya\\\AND Chala Fufa\\\And Charles Lee\AND Microsoft\\
  {\tt \{visuryan, bapatra, pamelabh, chfufa, charlle\}@microsoft.com}
  }

\date{}

\begin{document}
\maketitle
\begin{abstract}
A prominent problem faced by conversational agents working with large documents (Eg: email-based assistants) is the frequent presence of information in the document that is  irrelevant to the assistant. This in turn makes it harder for the agent to accurately detect intents, extract entities relevant to those intents and perform the desired action. To address this issue we present a neural model for scoping relevant information for the agent from a large document. We show that when used as the first step in a popularly used email-based assistant for helping users schedule meetings\footnote{We use Hedwig in lieu of the actual persona of the agent throughout this paper}, our proposed model helps improve the performance of the intent detection and entity extraction tasks required by the agent for correctly scheduling meetings: across a suite of 6 downstream tasks, by using our proposed method, we observe an average gain of 35\% in precision without any drop in recall. Additionally, we demonstrate that the same approach can be used for component level analysis in large documents, such as signature block identification.
\end{abstract}
\section{Introduction}
\label{sec:intro}

\blfootnote{
    %
    %
    \hspace{-0.65cm}  
    %
    %
    %
    %
    \hspace{-0.65cm}  
    This work is licensed under a Creative Commons 
    Attribution 4.0 International License.
    License details:
    \url{http://creativecommons.org/licenses/by/4.0/}.
}

\begin{figure*}[!htb]
    \centering
    \subfloat[A typical email encountered by the Scheduling Assistant]{
        \includegraphics[width=0.55\textwidth]{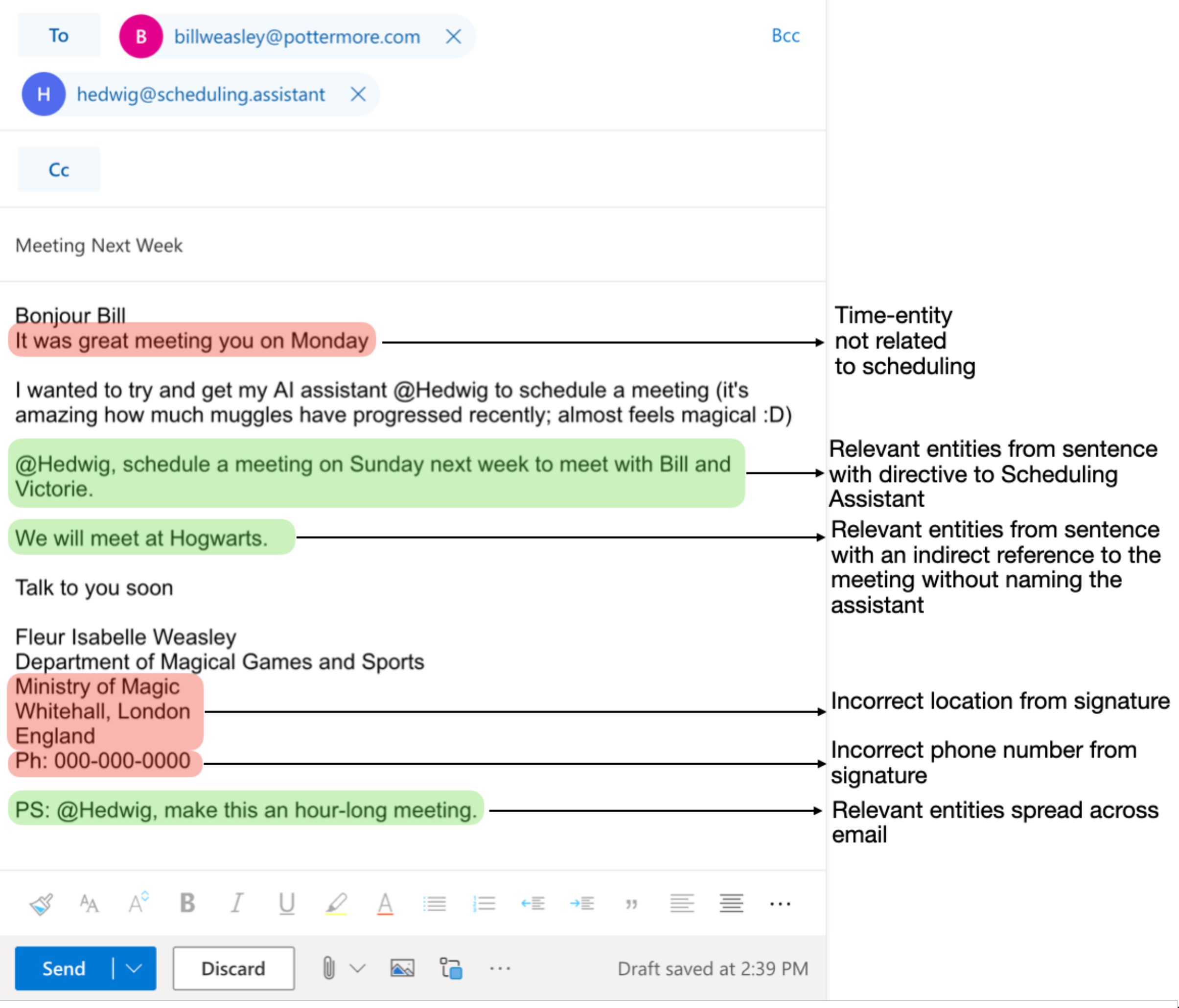}\label{fig:example}
    }
    \subfloat[The Email processing workflow]{
        \includegraphics[width=0.42\textwidth]{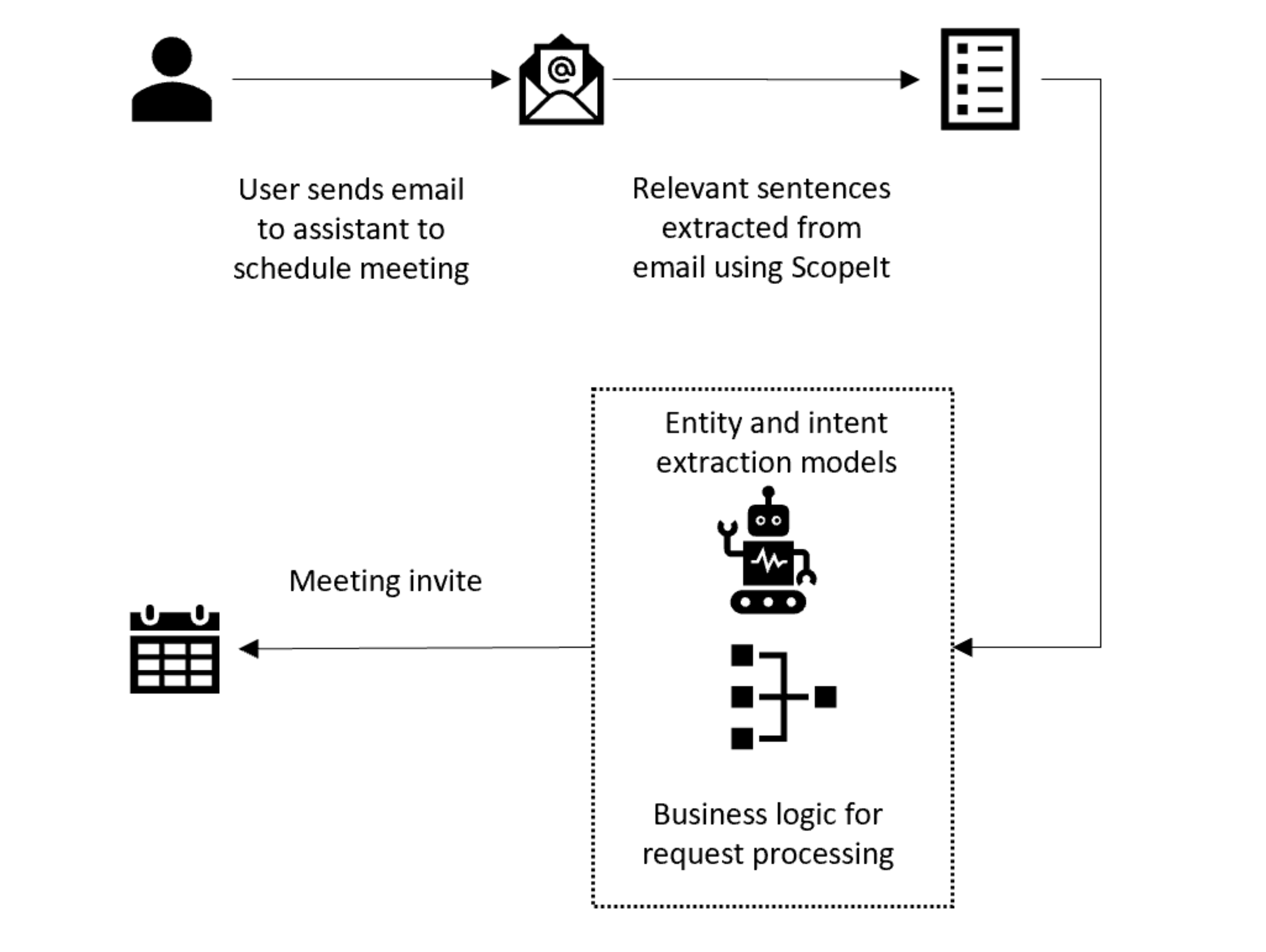}\label{fig:scopeitArch}
    }
\end{figure*}

Intelligent personal digital assistants (IPDA) such as Microsoft Cortana, Amazon Alexa, Apple Siri, and Google Assistant, are becoming increasingly popular. These assistants make use of natural language to communicate, which leads to faster task completion, improving the user’s productivity. A typical interaction with such a digital assistant requires a trigger (such as saying the assistant’s name), followed by a short phrase or sentence describing the user’s ask of the digital assistant. Some examples of these conversations are: \textit{``Cortana, what is the weather now?''}, \textit{``Alexa, play next''}, \textit{``Siri, turn off Bluetooth''}, \textit{``Hey Google, take me home''}. 

While most such assistants are voice based and communicate synchronously with the user, working mostly with short, targeted directives; there also exist email-based assistants that communicate and provide assistance asynchronously and thus have to work with much larger textual queries. Notable examples from the scheduling space are assistants like Cortana from Microsoft Scheduler, Amy and Andrew from x.ai, and Clara from Clara labs. These assistants require that the meeting organizer add them in the email with the attendees, and delegate the scheduling task to the assistant. Fig \ref{fig:example} shows an example of an email that an organizer can send to their virtual assistant. After receiving the email, the assistant needs to identify intents and entities of interest for scheduling the meeting correctly. For example, the duration of the meeting, where the meeting is (location), required and optional attendees, the type of meeting being requested (e.g. lunch, coffee), etc. This intent detection and entity extraction from large documents (eg: emails) can be challenging for two reasons:

\begin{itemize} 
\addtolength\itemsep{-2mm}
\item Information for scheduling the meeting could be spread across the document where most of the content is irrelevant
\item Most generic open source entity extraction models are recall-heavy as they often are context independent, and consequently detect entities that are not relevant for scheduling. 
\end{itemize}

Both the issues can be mitigated by building models (feature-based and/or neural) trained on the task at hand. However as we show in \S \ref{section:downstream}, these models can still get confused by the irrelevant information in the document and their performance can be improved by identifying relevant sentences of the document.

We model this problem of finding relevant sentences in a large document as a sentence-level binary classification problem, where every sentence in the email is either considered to be relevant or irrelevant to the context of scheduling meetings. While we focus on scheduling as an example throughout this paper, we believe our approach would be useful for domains outside of scheduling. We show that when used as a preprocessing step (Fig \ref{fig:scopeitArch}), a good performance of our proposed model (ScopeIt) on the task of identifying relevant sentences in an email boosts the performance of the downstream intent classifiers and entity extractors. Additionally, we show the utility of the same model for signature block detection in component level analyses of emails. We demonstrate that our method can identify signature blocks for signature removal tasks, often required for pre-processing emails for text to speech systems, or for anonymizing email corpora.

The main contributions of this work are:
\begin{itemize} 
\addtolength\itemsep{-2mm}
\item We propose a novel model (ScopeIt) for scoping out task relevant sentences from a large document that outperforms strong baseline methods
\item We illustrate the benefits of using ScopeIt as a preprocessing step and show that it improves the performance of a suite of downstream intent classifiers and entity detectors for the meeting scheduling task; improving precision by 35\% (average) without any drop in recall. To the best of our knowledge, this is the first work to explore the utility of scoping task relevant sentences as a preprocessing step for tackling problems involving large text corpora.
\item We show that our proposed architecture also performs better than publicly available baselines on the component level tasks like signature detection and generalizes better to real world data.
\end{itemize}

We present our approach to the problem of scoping out relevant sentences in \S \ref{section:approach}. In \S \ref{sec:experiments}, we describe our experimental setup and introduce the baselines we compare our approach against. We discuss ScopeIt’s performance in \S \ref{section:Results}. We analyze the embedding space induced by ScopeIt in \S \ref{section:latentSpace} to understand why it performs well. In \S \ref{section:downstream} we show the effectiveness of using ScopeIt as a preprocessing step on downstream intent classification and entity extraction tasks. We then show the performance of ScopeIt on the signature detection task (\S \ref{section:SigRemove}). In \S \ref{section:relatedWork} we discuss the related work. Finally, we conclude in \S \ref{section:conclusion}.

\section{Proposed Method}
\label{section:approach}
\begin{figure*}[!thb]
\centering
\includegraphics[width=0.80\textwidth]{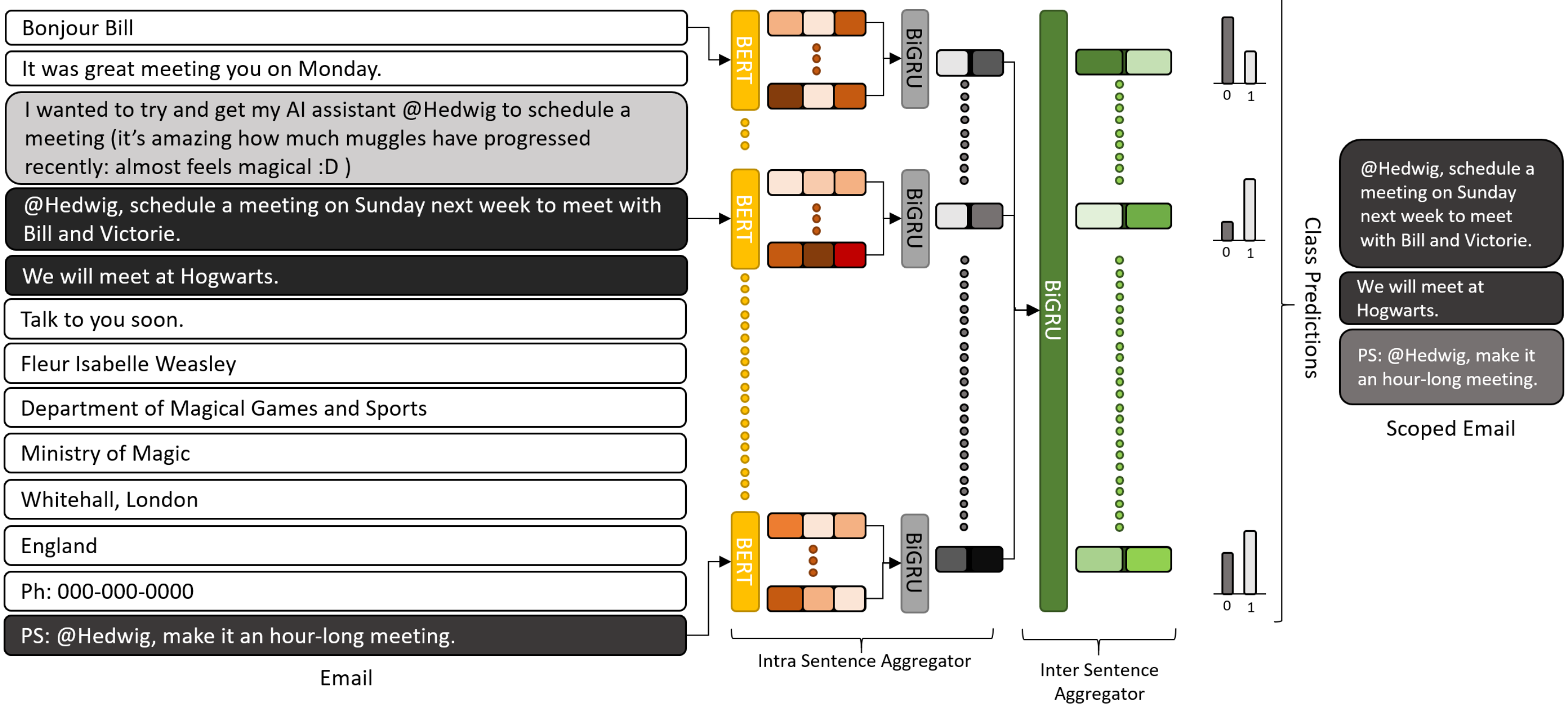}
\caption{The model architecture: Contextual embeddings for each token in a sentence are first generated using BERT, following which, a BiGRU is used to generate the sentence embeddings. A BiGRU then aggregates information across the document. Finally, we predict the probability of selecting a sentence.}
\label{fig:model} 
\end{figure*}

In this section, we outline our approach to the problem of scoping out relevant sentences for a NLP-based scheduling assistant. Our approach consists of 2 parts: a preprocessing module and a neural model. An incoming email is first passed through the preprocessing module. The preprocessed email is then tokenized, indexed and passed through the neural model to generate a confidence score for each sentence. The model is trained end-to-end with human-labeled gold scores denoting the relevant sentences of the email. We also adopt some data augmentation methods to improve model generalization.

\subsection{Preprocessing Module}
\label{sec:preprocessing}
The preprocessing step fixes any issues due to improperly decoded text (mojibake characters). Furthermore, since we use the wordpiece tokenizer\footnote{\url{https://github.com/google/sentencepiece}} to tokenize each word into its constituent wordpieces, having raw urls or emails often generates a large number of uninformative wordpieces\footnote{E.g ``https://coling2020.org/pages/call\_for\_papers.html'' generates 20 wordpieces}. In order to circumvent this issue, we replace all urls and emails with special tokens (eg: URLTOKEN, EMAILTOKEN). We keep track of the original urls/emails, and invert the token replacement after obtaining the confidence scores from the neural model.

\subsection{Neural Model}
\label{sec:model}
Our neural model consists of 3 different modules: an intra-sentence aggregator to aggregate information within a sentence, an inter-sentence  aggregator to share information across different sentences, and a classifier to predict the final relevance score of each sentence (Fig \ref{fig:model}). 
Given a document, we first tokenize it into sentences using NLTK's sentence tokenizer. We then use the wordpiece tokenizer to tokenize each sentence. Let $\mathcal{X} = \{w_{1, 1} \cdots w_{m, l_m}\}$ be the tokenized document, where $w_{i,j}$ denotes the $j^{th}$ wordpiece of the $i^{th}$ sentence and $l_m$ denotes the length of the $m^{th}$ sentence. We predict the relevance of each sentence using the following approach:

\begin{minipage}{0.50\textwidth}
    \centering
    
        \begin{equation*}
            \begin{aligned}
            (e_{i,1}, \cdots e_{i, l_{i}}) &= BERT(w_{i,1} \cdots w_{i, l_i}) \\
            ({h_f}_{i,1} \cdots {h_f}_{i, l_i}) &= \overrightarrow{Seq2SeqEncoder}(e_{i,1}, \cdots e_{i, l_{i}}) \\
            ({h_b}_{i,1} \cdots {h_b}_{i, l_i}) &= \overleftarrow{Seq2SeqEncoder}(e_{i,1}, \cdots e_{i, l_{i}}) \\
            e_{s_i} &= [{h_f}_{i,l_i}; {h_b}_{i,1}]
            \end{aligned}
            \end{equation*}
    \end{minipage}
    \begin{minipage}{0.47\textwidth}
        \centering
        \begin{equation*}
            \begin{aligned}
            (f_{s_1}, \cdots f_{s_m}) &= \overleftrightarrow{Seq2SeqEncoder}(e_{s_1}, \cdots e_{s_m}) \\
            (p_{s_1}, \cdots p_{s_m}) &= (\sigma(f_{s_1}), \sigma(f_{s_2}) \cdots \sigma(f_{s_m}))
            \end{aligned}
        \end{equation*}
\end{minipage}

{\bf Intra Sentence Aggregator:} Let $s_i = \{w_{i, 1} \cdots w_{i, l_i}\}$ be the $i^{th}$ sentence. We generate contextual embeddings for each token $w_{i,j} (1\leq j\leq l_i)$. We use BERT \cite{bert} for generating the embeddings. Note that generating embeddings for each sentence independently, along with replacing urls with special tokens, allows us to circumvent the issue of BERT having a maximum of 512 positional embeddings (i.e we can now encode 512 * $num\_sentences$ wordpieces). Since we want to avoid back-propagating through BERT for compute constraints, using the [CLS] token (as is commonly done to generate sentence representations) doesn't work. Consequently, we use a Seq2Seq encoder to better adapt the contextual embeddings to the task. We then concatenate the final forward and backward hidden dimensions to get the sentence embedding $e_{s_i}, 1 \leq i \leq m$ for each sentence:

{\bf Inter Sentence Aggregator:} Given sentence embeddings $\{e_{s_1} \cdots e_{s_m}\}$, we use a Seq2Seq encoder to aggregate information across different sentences. This allows the model to capture context based on other sentences around it, enabling us to capture document level features. A final Sigmoid output layer generates the probability of each sentence being relevant:

The model is trained with a binary cross entropy loss using gold annotated relevance scores, i.e. given annotations for the sentences $\mathcal{Y} = \{y_1, y_2, \cdots y_m\}$:
\begin{equation}
\begin{aligned}
\mathcal{L} = -\sum_{i=1}^{m} y_{i} \log(p_{s_i}) + (1 - y_{i}) \log(1 - p_{s_i})
\end{aligned}
\end{equation}

\subsection{Data Augmentation}
\label{sec:dataAug}

Given that most emails received by the scheduling assistant have some information pertinent to scheduling, we augment the training data with irrelevant emails (i.e emails not relevant to scheduling). These emails are sampled from the Enron dataset \cite{enron,klimt2004introducing}. Further, we observed that the model was confused when given texts that did not resemble general email writing styles. To avoid this bias, we augment the dataset with negative samples from the IMDb and Yelp datasets \cite{imdb-yelp}. Furthermore, we observed that the original dataset had a bias of having relevant information being present at the beginning of the email. In order to account for that bias, we also shuffled passages of text within each email except the salutation and signature, and augment our dataset with the shuffled emails. We do so to ensure that the resulting shuffled emails are not nonsensical. We also augment the dataset by first creating templates for emails that are representative of the emails the system would receive, and then randomly replacing proper nouns in the email templates. Additional details can be found in Appendix \ref{app:datasetDetails}.

\section{Experiments on Scoping Relevant Sentences}
\label{sec:experiments}
\label{sec:RelevanceScoping}

\subsection{Dataset and Experimental Setup}
We show the performance of ScopeIt on an internal dataset for identifying relevant sentences from emails for the context of scheduling. The details of pertaining to the dataset creation can be found in Appendix \ref{app:datasetDetails}. Table \ref{tab:relevanceScopeDataset} shows the instances present in the train, validation and test splits. During evaluation, any sentence with score $>0.5$ is classified as relevant and others are classified as irrelevant. We use the F1 score of the sentence relevance prediction task as the metric of evaluation. We report the hyperparameters and training details in Appendix \ref{app:hyperparams}.

\subsection{Baselines}

\paragraph{Seq2Seq Encoder:} This model does not use BERT for generating contextual embeddings. Instead, a standard word-level BiGRU model is used as the sentence encoder to generate sentence embeddings, with the vocabulary set to the top 10,000 most frequently occurring words encountered in the training data. The sentence embeddings are then projected using a feed-forward layer to generate the relevance probabilities.

\paragraph{No Inter-Sentence Aggregator:} This model uses BERT for generating the contextual embeddings, and then a BiGRU encoder to generate the sentence embeddings. It however does not make use of any inter-sentence aggregator; instead a feed-forward layer directly generates the relevance probabilities.

\paragraph{BERT with [CLS] only:} This model just uses the [CLS] token of BERT for generating the sentence embedding vector. Note that we don't fine-tune the BERT model.

\paragraph{ScopeIt Without Data Augmentation:} Our proposed model without any data augmentation (\S \ref{sec:dataAug})

\paragraph{BertSum:} A state of the art extractive summarization model leveraging BERT \cite{liu2019text}. We use the implementation provided by the authors.

\begin{table}[!htb]
    \small
\begin{minipage}{0.50\textwidth}
    \centering
    \begin{tabular}{@{}l@{}r@{}r@{}r@{}r@{}r@{}}
    \toprule
    {{\bf Split} }  & { {\bf n\_docs} } & { {\bf n\_internal} } & { {\bf n\_sent} } & { {\bf n\_pos} } & { {\bf n\_neg}} \\
    \midrule
    train    & { 21875 }   & { 10546 } & { 233307 } & { 24428 } & { 208879 } \\
    \midrule
    validate & { 2436 }  & { 1176 } & { 25866 } & { 2699 } & { 23167 } \\
    \midrule
    test\footnotemark     & { 1215 } & { 1015 } & { 12055 } & { 1716 } & { 10339 } \\
    \bottomrule
    \end{tabular}
    \caption{Relevance Scoping Dataset Details}
    \label{tab:relevanceScopeDataset}
\end{minipage}
\begin{minipage}{0.5\textwidth}
    \centering
    \begin{tabular}{@{}l@{}r@{}}
    \toprule
    {\bf Model} & {\bf F1 Score} \\
    \midrule
    BERT [CLS] & 0.81 \\
    Seq2Seq Encoder & 0.83 \\
    No Inter-Sentence Aggregator & 0.89 \\
    BertSum \cite{liu2019text} & 0.90 \\
    ScopeIt Without Data Augmentation & 0.93 \\
    ScopeIt & {\bf 0.94} \\
    \bottomrule
\end{tabular}
    \caption{Performance for Relevance Scoping}
    \label{tab:relevanceScope}
\end{minipage}
\end{table}
\footnotetext{We augment the test set with 200 completely irrelevant emails to gauge the model performance for that scenario.}

\vspace*{-4mm}
\section{Main Results}
\label{section:Results}
Table \ref{tab:relevanceScope} shows the performance of ScopeIt compared to the baseline models. Since the BERT [CLS] model is not fine-tuned, it does not perform as well as any of the models where the Seq2Seq encoders are trained. Unsurprisingly, the models with BERT augmented embeddings outperform the standard Seq2Seq encoder model substantially. We observe that the inter-sentence aggregator also improves performance. Finally, the model with data-augmentation outperforms all of the baselines. We believe this is because of two reasons. First, most emails have a prior of being relevant, simply because the user cc'd the scheduling assistant. Consequently, the model predicted some sentences as relevant, even for emails which did not have any. Augmenting the data with completely irrelevant emails helps overcome that bias. Further, for most emails, the relevant scope occurs in the beginning of the email. Hence, baseline models bias towards scoring the beginning of the email higher than the end, even if the beginning was not particularly relevant. Training with the shuffling data augmentation mitigates the issue.

Our proposed model also performs better than BertSum. We hypothesize this is because our dataset is orders of magnitude smaller than the CNN/Daily Mail dataset \cite{hermann2015teaching} used by \newcite{liu2019text}. And while finetuning BERT models for general tasks does not require as much data, BertSum uses a formulation very different from the original BERT model \cite{bert}\footnote{Specifically, they use alternating type tokens for each sentence, with each sentence separated with a [SEP] token and a [CLS] token, and use the [CLS] token of the sentence to generate the sentence embedding. }. Consequently, finetuning BERT to adapt to these modification potentially requires more data. Moreover, BertSum still suffers from the 512 wordpiece restriction (\S \ref{sec:preprocessing}), while ScopeIt does not.
\section{Clustering in the Embedding Space}
\label{section:latentSpace}
We next investigate if sentence embeddings generated by ScopeIt exhibit any clusters that make semantic sense. On preliminary analysis, some clusters that we observed in the data were salutations, signature blocks and sentences containing entities associated with scheduling meetings. We hypothesize that similar clusters should be observed in the sentence embedding space. To test this hypothesis, we propose the following experiment: given the embedding of a sentence belonging to a certain cluster (the query sentence), retrieve the top k nearest neighbor sentence embeddings from a set of sentence embeddings generated by ScopeIt. If similar clusters exist in the embedding space, then sentences associated with the retrieved embeddings should belong in the same cluster as the query sentence.

Due to space constraints, we describe the methodology of the embedding experiment in Appendix \ref{app:embeddingSpace}, and report our main findings here. We observe that salutations and signatures are clustered together. We also observe sub-clusters wherein sentences containing similar entities or intents are clustered with sentences containing similar information. Moreover, we find that these sentence embeddings also capture the context in which the sentences occur: syntactically similar query sentences get mapped to different clusters based on the context in which they occur.

\section{Improvements to Downstream Tasks}
\label{section:downstream}

Our main motivation for developing ScopeIt was the hypothesis that using relevant sentences in place of the entire document would improve the performance of downstream NLP tasks. In this section we highlight the impact ScopeIt has on $6$ downstream tasks, $5$ of which are either associated with detecting an intent related with scheduling a meeting or extracting the necessary entities. The models for tackling these tasks use the scoped message generated by ScopeIt as the input. We also consider the ``Non-actionable Emails'' task which helps the scheduling assistant identify emails that it should ignore. The models used for each task vary: they can either be context independent regex models, or context aware neural models. For each of these tasks, we first describe what the task is, and then describe the model(s) used for solving it. Finally, to show the impact of ScopeIt, for each task, we give the model(s) the original unaltered email and the scoped version as input, and compare the performance difference. We summarize the results of these experiments in \S \ref{sec:summary}. Note that there is no overlap between the data used for the analysis presented in this section and the data used for training ScopeIt.

\paragraph{1. Meeting Type} When the scheduling assistant receives an email and has determined that the email has an intent to schedule a meeting, the Meeting Type task tries to classify the meeting request into one of the broad classes of meeting types as defined by the system. Each of the categories have special meeting properties that help the assistant populate the meeting details. Some examples of these meeting types defined are Lunch (which constrain the times to schedule), Conference Call (require a Remote Bridge), Phone call etc. The assistant uses an ensemble of different models to classify the meeting requests into these classes. For this case study, we focus on the model responsible for detecting a call or a conference call intent, which maps to the Phone Call and Conference Call meeting type classes, respectively.

\indent { \bf Example Input:} {\it``Let us get together on a Team's call.''} \\
\indent { \bf Expected Output:} Conference Call Intent

This task is modeled as a multi-label classification task, and we use a context aware deep network to tackle it. We use a model similar to the one proposed in \cite{mullenbach-etal-2018-explainable}. Specifically, we generate a contextual embedding using BERT for each token in the email. Then, an attention method \cite{bahdanau2014neural}, one for each label, is used to aggregate the embeddings into a document embedding, which is then passed through a sigmoid layer to generate the probability for each label. The entire model is trained end to end by minimizing the negative log likelihood of the gold labels. While using ScopeIt, we only select sentences that occur above a particular threshold (0.01), and feed the concatenation of those sentences as inputs to the model.

\paragraph{2. Meeting Duration} The scheduling assistant needs to extract the duration for the meeting from the meeting organizer's email. If there wasn’t a duration entity detected, the system uses the default duration set by the organizer in their meeting preferences.

\indent {\bf Example Input:} {\it ``Hedwig, schedule a meeting for 30 minutes.''} \\
\indent {\bf Expected Output:} 30 minutes.

We use LUIS\footnote{\url{https://www.luis.ai/home}} for extracting the duration of a meeting from the meeting requests. LUIS is the Language Understanding Service in Microsoft Azure Cognitive Services that provides natural language intelligence for conversational AI applications \cite{williams2015fast}. In order to utilize LUIS’ high recall duration extraction model in the context of scheduling meetings, we select sentences scored above 0.01 by ScopeIt, and feed the concatenation of the sentences as the input to LUIS’ duration extraction model.

\paragraph{3. Meeting Phone Number} When users schedule a phone call, the system needs to extract phone numbers from the organizer or attendee to add to the meeting invite.

\indent {\bf Example Input:} {\it ``Hedwig, please schedule a call with Albus. My phone number is +1 000-000-0000. Regards, Gellert Grindelwald''} \\
\indent {\bf Expected Output:} +1 000-000-0000.

We use LUIS for extracting the phone numbers from an email. We extract sentences scored above a threshold of 0.01 by ScopeIt, concatenate the sentences, and feed that as an input to the high recall phone number extraction model.

\paragraph{4. Meeting Location} In order to schedule the meeting at the right location, the system needs to extract the intended location expressed by the organizer.

\indent {\bf Example Input:} {\it ``Hedwig, schedule a meeting. Hagrid, let's meet at the 3 Broomsticks.''} \\
\indent {\bf Expected Output:} the 3 Broomsticks

This is modeled as an entity extraction problem and consequently we fine-tune BERT for tagging (similar to the BERT for NER, as done in \newcite{bert}). We concatenate sentences scored above a certain threshold by ScopeIt and pass it as an input to the model.

\paragraph{5. Meeting Timezone} Users typically express multi timezones in two ways: express time zones by explicitly mentioning timezone abbreviations like "EST", or implicitly by indicating the city and sometimes the country where the meeting is going to be held.

\indent {\bf Example Input:} {\it ``Hedwig, schedule an online meeting with Ron Weasley next week. Ron is in EST, and I am going to be working from Dublin for that week.''} \\
\indent {\bf Expected Output:} EST, Dublin

By using ScopeIt to filter out sentences irrelevant to scheduling the meeting, the system is able to leverage recall-heavy time zone entity extractors, and city and country extractors to find the right time zones.  It utilizes LUIS for time zone entity extraction and LU (Location Understanding) from Bing to extract cities and countries from the input text. These utterances are subsequently resolved for their time zone offsets.

\paragraph{6. Non-actionable Emails} When the scheduling assistant processes a request, the system might receive emails from meeting participants which are irrelevant to scheduling. For example, after the meeting organizer has sent a request to the scheduling assistant (Hedwig, in the prior examples), one of the invitees might reply to the email thread with all meeting participants including the assistant saying, {\it ``Thanks for setting this up. Look forward to meeting you.''} In these cases, there is no action required from the system's point of view and the email can be safely ignored. Similar to the approach stated in the previous tasks, sentences in the email that are scored above a threshold are extracted and concatenated. If there are no sentences in the email above the relevance threshold, the email is considered irrelevant and is ignored by the system.

\subsection{Results}
\label{sec:summary}
\begin{table*}[!thb]
\small
\centering
\begin{tabular}{@{}lcccccr@{}}
\toprule
{\bf Task} & {\bf Task Type} & {\bf Model Type} & {\bf Metric} & {\bf Before ScopeIt} & {\bf After ScopeIt} & {\bf $\Delta$} \\
\midrule
Meeting Type & Classification & Context Aware & Accuracy & 0.72 & 0.96 & +0.24 \\
Non-actionable Emails & Classification & Context Aware & Accuracy & N/A & 0.96 & + 0.96 \\
\midrule
\midrule
Duration & Extraction & Context Independent & Accuracy & 0.88 & 0.92 & +0.04 \\
\midrule
\multirow{2}{*}{Phone Number} & \multirow{2}{*}{Extraction} & \multirow{2}{*}{Context Independent} & Precision & 0.46 & 0.98 & +0.52 \\
    & & & Recall & 1 & 1 & 0 \\
\midrule
\multirow{2}{*}{Location} & \multirow{2}{*}{Extraction} & \multirow{2}{*}{Context Aware} & Precision  & 0.73  & 0.96 & +0.23 \\
& & & Recall & 0.92 & 0.96 & +0.04 \\
\midrule
\multirow{2}{*}{Timezone} & \multirow{2}{*}{Extraction} & \multirow{2}{*}{Context Independent} & Precision & 0.37 & 0.67 & +0.30 \\
    &  &  & Recall & 0.92 & 0.96 & +0.04 \\
\midrule
\midrule
\multicolumn{3}{c}{\multirow{3}{*}{Average}} & Accuracy & & & +0.14 \\
 & & & Precision & & & +0.35 \\
 & & & Recall & & & +0.02 \\
\bottomrule
\end{tabular}
\caption{A summary of all improvements resulting from the ScopeIt's preprocessing} 
\label{tab:downstream-summary}
\vspace{-2mm}
\end{table*}
Table \ref{tab:downstream-summary} summarizes the utility of using ScopeIt. For the intent classification and duration extraction tasks, we see an average increment of 0.14 in the accuracy. An interesting observation is that even the context aware neural model benefits strongly (+0.24 accuracy improvement).

For the entity extraction models, we observe a strong increase in precision, with an average increase of 0.35. The context independent models benefit strongly when we strip out the irrelevant parts of the document: as shown in the example in Figure \ref{fig:example}, phone numbers extracted by the context independent regex based model are often found in the signature block of the email. A similar behavior is also observed in the timezone extraction task, where locations in the signature often get picked up as timezones. As hypothesized, once the email is scoped to only the relevant parts, these models get a substantial boost in precision. A similar gain is also observed for the BERT Location extractor. 

An interesting observation is that the recall for these extraction models also improves. On further investigation, we found that this can be attributed to an increase in the true positives. For the BERT Location extraction, this makes sense, since a simplified input allows the model to reason better about the location. For the timezone task, we hypothesize that the LU model has additional heuristics and that the heuristics perform better on the simplified inputs.

Using ScopeIt also offers the benefit of making regex models feasible to use. This is especially advantageous since regex-based models have faster inference times and require much less data to build than their neural counterparts.

Finally, ScopeIt also helps the scheduling assistant decide between which emails to process and which ones to ignore, which plays a crucial role for an email-based agent. People often use reply-all while interacting with each other on an email thread. This leads to the agent receiving emails whose contents are not relevant to the task of scheduling the meeting. Using ScopeIt ensures that those emails are ignored by the agent.

\section{Signature Block Detection}
\label{section:SigRemove}
As described in \S \ref{section:latentSpace}, we observed that sentences with similar semantics were clustered close to each other in the sentence embedding space.  We use this observation to apply our model to component detection in email, specifically for signature block identification. We show the model's performance on a publicly available dataset, and show that it outperforms the baseline model. We also hypothesize that the publicly available systems for extracting signatures are not suitable for real-world use-cases, as they are often trained on well structured emails using hand crafted features, and hence are not robust to the variety of writing styles that people employ in the real world. In order to validate this hypothesis, we test the effectiveness of the baseline on our use-case.

\subsection{Dataset and Experimental Setup}
We use the 20-Newsgroup dataset consisting of emails annotated with signature blocks \cite{carvalho04ceas}. This dataset is publicly available \footnote{\url{https://www.cs.cmu.edu/~vitor/codeAndData.html}}. We use a standard split of 80\%, 10\% and 10\% as the training, validation, and testing splits. In order to validate our hypothesis about the efficacy of the publicly available baseline on our use-case, we annotate 625 emails with signature blocks and then test the performance of the baseline as well as our model (trained on the 20-Newsgroup dataset) on this annotated dataset. The number of instances in the dataset can be found in Table \ref{tab:JangadaVsScopeItDatasetDetails}.

\subsection{Baseline}
We compare against a publicly available signature detection tool Jangada\cite{carvalho04ceas}. Jangada uses a CRF model with handcrafted features, and is trained on the 20-Newsgroup dataset.

\begin{table}[!thb]
    \small
    \centering
    \begin{minipage}{0.48\textwidth}
        \begin{tabular}{@{}lcrrrr@{}}
    \toprule
    {\bf Dataset} & {\bf Split} & {\bf n\_docs} & {\bf n\_sent} & {\bf n\_pos} & {\bf n\_neg} \\
    \midrule
    \multirow{3}{*}{\makecell[c]{20 \\ Newsgroup}} & train & 465 & 20629 & 18076 & 2553 \\
    \cmidrule{2-6}
    & val & 52 & 1797 & 1522 & 275 \\ 
    \cmidrule{2-6}
    & test & 100 & 4547 & 4054 & 493 \\
    
    \midrule
    \multirow{3}{*}{\makecell[c]{Manually \\ Annotated}} & train & 501 & 6055 & 2043 & 4012 \\
    \cmidrule{2-6}
     & val & 62 & 670 & 227 & 443 \\
    \cmidrule{2-6}
    & test & 62 & 663 & 260 & 403 \\
    \bottomrule
\end{tabular}
        \caption{Signature Detection: Dataset Details}
        \label{tab:JangadaVsScopeItDatasetDetails}
        \vspace{-2mm}
    \end{minipage}
    \begin{minipage}{0.48\textwidth}
        \begin{tabular}{@{}lcrrr@{}}
    \toprule
    {\bf Dataset} & {\bf Model} & {\bf Precision} & {\bf Recall} & {\bf Fscore}  \\
    \midrule
    \multirow{2}{*}{20} & {Jangada} & 0.98 & 0.971 & 0.975 \\
    \cmidrule{2-5}
    Newsgroup & {ScopeIt} & 0.992 & 0.999 & {\bf 0.996} \\
    \midrule
    \multirow{2}{*}{Manually} & {Jangada} & 0.908 & 0.224 & 0.359 \\
    \cmidrule{2-5}
    Annotated & {ScopeIt} & 0.995 & 0.884 & {\bf 0.936} \\
    \bottomrule
\end{tabular}
        \caption{Performance: Jangada Vs ScopeIt}
        \label{tab:JangadaVsScopeIt}
        \vspace{-2mm}
    \end{minipage}

\end{table}

\subsection{Results on Signature Block Detection}
As seen in Table \ref{tab:JangadaVsScopeIt}, our proposed neural model outperforms Jangada on the 20 Newsgroup dataset. We also validate our hypothesis: when we use Jangada for our real world use-case to remove signatures, we observe that while it has a high precision, the recall drops drastically (0.224); making it impractical to use in production. On the other hand, ScopeIt, even when trained on 20 Newsgroup, generalizes much better (recall 0.885, fscore: 0.936).
\section{Related Work}
\label{section:relatedWork}

Our problem of relevance scoping in documents is similar to extractive summarization. Extractive summarization deals with selecting subsets (usually sentences) of a document that succinctly summarizes it. For the case of this scheduling assistant, scoping out the relevant part in an email document is in essence selecting the subset of sentences from the email that accurately summarizes the scheduling intent and specifies the parameters necessary to schedule a meeting correctly. Both traditional feature based methods using word probability, TF-IDF weights, sentence position and sentence length features \cite{luhn1958automatic,eduard1998automated,cao2015learning,ren2016redundancy} and recent neural methods \cite{nallapati2017summarunner,zhou2018neural,narayan2018:ranking,liu2019:single,liu2019text} have been used for the task of extractive summarization. \newcite{liu2019text} show the benefit of using pretrained language models \cite{elmo,radford2018improving,bert,dong2019unified,hibert} for the same task. Their proposed model BertSum leverages interval segment embeddings to distinguish multiple sentences within a document. BertSum further also finetunes the BERT embeddings to learn the segment embeddings during training, which potentially requires more data; and can also only encode upto 512 wordpiece long documents. In contrast, we used hierarchical RNNs (similar to \cite{nallapati2017summarunner,zhou2018neural}), with the pretrained embeddings forming the embedding layer (Fig. \ref{fig:model}); thereby allowing us to encode emails much larger than 512 tokens long.

Intent classification and entity extraction tasks in the context of conversational understanding have been studied both in academia and corporate research laboratories \cite{tur_spokenLanguage}. There exists a rich body of research in user intent identification from targeted queries \cite{wang_emnlp2014}. However, these methods don't work as well when applied to large documents. We showed that scoping out the relevant parts in a document improves performance of classification and extraction tasks on large queries. To the best of our knowledge, this is the first work to explore the utility of extractive summarization as a preprocessing step for tackling problems involving large text corpora.

There has been extensive research on the topic of identifying signature blocks and reply lines from an email \cite{carvalho04ceas,minkov-etal-2005-extracting,balog2006finding,xiaoqin2015unsupervised}. \cite{balog2006finding,xiaoqin2015unsupervised} present heuristic driven methods for unsupervised identification of the signature body, while \cite{carvalho04ceas} present a CRF based approach for identifying and extracting signature and reply lines from Email. We showed in \S \ref{section:SigRemove} that our proposed method also works well for removing signatures and also generalizes better.

\section{Conclusion}
\label{section:conclusion}
In this paper, we proposed a simple method for scoping relevant information within emails and the impact the model had on a suite of tasks that are vital for the scheduling assistant. We also showed our models applicability on the task of Signature Detection. We show that it performs better than existing publicly available baselines and generalizes better on real world use-cases.

In this work we showed that our model works well with emails. For future work, we plan on investigating the impact of our proposed method for other tasks that process large textual inputs (Eg: document classification, sentiment analysis on large reviews). Furthermore, using BERT for inference poses latency challenges in a production system. A promising direction of future work that we plan on investigating is leveraging distilled versions of BERT \cite{distilBERT,wang2020minilm} for the task.

\bibliographystyle{coling}
\bibliography{coling2020}
\clearpage
\appendix
\section{Dataset Creation Details}
\label{app:datasetDetails}
The dataset is built by sampling 12737 emails from an internal dataset. These emails were split into a train and test set first at a 90\%-10\% split, and then the train set was split again at a 90\%-10\% split to form the training and validation datasets. In order to measure the inter-annotator agreement for the dataset, we randomly sample 200 emails, which were then annotated by another annotator. The Cohen's kappa ($\kappa$) measured was 0.89.

This dataset is augmented with negative samples from the Enron dataset, which are used to account for emails from a professional settings, which are not related to scheduling. This is done by using a list of disqualification words that remove emails that could potentially be referring to meetings. The disqualification words for Enron are as follows: ``book a room'', ``let's meet'', ``meeting'', ``conference room'', ``meet'', ``invitation'', ``location'', ``half an hour'', ``30 mins'', ``30 minutes'', ``45 mins'', ``schedule'' and ``reserve''. If any of these phrases are found in the email, the same disqualifies them as a negative. A total of 5429 emails were added from the Enron dataset.

To account for emails that do not conform to regular language used in a professional setting, text data was sampled and added from the Yelp and IMDB subsets from the UCI Sentiment Labelled Sentences dataset. Specifically,  1000 documents and 748 documents were sampled from the Yelp and IMDB subsets respectively, after similar disqualification rules. These were added to the train and validation dataset in the same proportions as before. The data augmentation was necessary to ensure the model is not biased to believe that all documents contain something relevant. We also add 200 examples to the test set to check if the trained models learn discard completely irrelevant documents.

Other data augmentation methods included random replacement and random swapping. For random replacement, proper nouns in templatized emails were replaced. Finally, the dataset was further augmented by random swapping of passages for longer emails (with $>3$ passages). The last and first passages were not swapped, as these generally tended to be salutation or signature blocks. These account for the remainder of the data points in the dataset.

\section{Hyperparameters and Training Details}
\label{app:hyperparams}
We use the BERT-Base, Multilingual Cased\footnote{\url{https://github.com/google-research/bert/blob/master/multilingual.md}} model for generating the contextual embeddings. We do not fine-tune the BERT model for any of the models due to compute constraints. We use a 2 layer BiGRU encoder \cite{gru} (hidden dimension of 128) as the Seq2SeqEncoder for the intra-sentence aggregator, and a 2 layer BiGRU (hidden dimension of 128) as the Seq2Seq Encoder for the inter-sentence aggregator. The model was trained with gradient descent for 50 epochs. We used Adam \cite{adam} as the optimizer with a learning rate of $0.0001$. The learning rate was annealed by a factor of 0.5 if the validation loss failed to improve over 5 epochs, and also use early stopping with a patience of 8. All our models were developed using the AllenNLP framework \cite{Gardner2017AllenNLP}.
For tuning the models, a grid search over the following learning rates was done: $\{1e-2, 5e-3, 1e-3\}$, as well as a batch size of $\{4, 8, 16, 32\}$ (beyond that caused out of memory issues). All models were trained on a single K80 instance.

\section{Clustering in Embedding Space}
\label{app:embeddingSpace}
\begin{table*}[!thb]
\footnotesize 
\centering
\begin{tabular}{p{3cm}p{2.5cm}p{1.5cm}p{7cm}}
\toprule
{\bf Original Email} & {\bf Query} & {\bf Cluster Type}  & {\bf NN With Surrounding Context} \\
\midrule
\multirow{9}{*}{\makecell[l]{Hey Harry\\\\\textcolor{red}{{\bf I'm using Hedwig}}\\\textcolor{red}{{\bf to schedule a}}\\\\\textcolor{red}{{\bf meeting! @Hedwig,}}\\\textcolor{red}{{\bf schedule a meeting}}\\\textcolor{red}{{\bf for next week,}}\\\textcolor{red}{{\bf in Hogsmade.}}\\\\Thanks,\\Ronald Weasley\\The Burrow\\Ottery St. Catchpole\\England}} & \multirow{3}{*}[-18pt]{Hey Harry} &  \multirow{3}{*}[-18pt]{Salutation} & \makecell[l]{\textcolor{blue}{{\it Hi Richard,}}\\Per my voicemail, are you available for w/Greg}\\
\cmidrule(){4-4} \\
 & & & \makecell[l]{\textcolor{blue}{{\it Jim,}}\\Is there going to be a conference call or some other\\ type of weekly meeting $\cdots$} \\
 \cmidrule{4-4} \\
 & & & \makecell[l]{\textcolor{blue}{{\it Hi Shirley,}}\\Is this meeting still set for tomorrow? $\cdots$} \\
 \cmidrule{2-4} \\
& \multirow{3}{*}[-24pt]{\makecell[l]{@Hedwig,\\schedule a\\meeting for\\next week,\\in Hogsmade.}} &  \multirow{3}{*}[-24pt]{\makecell[l]{Date-time\\availab-\\-ility\\intent}} & \makecell[l]{$\cdots$ call memo that we will forward on early next week. \\\textcolor{blue}{{\it Chris Long will be in touch on Tuesday to help}}\\\textcolor{blue}{{\it coordinate the recommended call.}}$\cdots$} \\
\cmidrule{4-4} \\
 &  & & \makecell[l]{$\cdots$ Any possibility of rescheduling to another day?\\\textcolor{blue}{{\it Sally is available Thursday, June 1.}} $\cdots$} \\
 \cmidrule{4-4} \\
 & & & \makecell[l]{Susan,\\\textcolor{blue}{{\it please organize a meeting with Steve, Kim, and Tracey}}\\\textcolor{blue}{{\it  early next week, say Monday or Tuesday}} $\cdots$} \\
 \cmidrule{2-4} \\
 & \multirow{3}{*}[-30pt]{Ronald Weasley} & \multirow{3}{*}[-30pt]{Signature} & \makecell[l]{$\cdots$ Thank you\\\textcolor{blue}{{\it Mona ********}}} \\
 \cmidrule{4-4} \\
 & & & \makecell[l]{$\cdots$ Thanks,\\\textcolor{blue}{{\it Larry ********}}} \\
 \cmidrule{4-4} \\
 & & & \makecell[l]{$\cdots$ Thanks,\\\textcolor{blue}{{\it Patti}}} \\
 \midrule
 \multirow{6}{*}{\makecell[l]{Hey Ron,\\\\Sounds's good.\\\textcolor{red}{{\bf Let's meet at the }}\\\textcolor{red}{{\bf Three Broomsticks.}}\\\textcolor{red}{{\bf @Hedwig, Ron}}\\\textcolor{red}{{\bf will call me.}}\\\textcolor{red}{{\bf  My phone number}}\\\textcolor{red}{{\bf is 000-000-0000.}}\\\\\\Thanks,\\Harry Potter\\Ph: 000-000-0000}} & \multirow{3}{*}[-35pt]{\makecell[l]{My phone number\\is 000-000-0000.}} & \multirow{3}{*}[-35pt]{\makecell[l]{Phone\\availability \\intent}} & \makecell[l]{$\cdots$ I should contact your assistant to schedule a meeting.\\\textcolor{blue}{{\it If you need to contact me immediately, please call my}}\\\textcolor{blue}{{\it cell phone at 000-000-0000.}} $\cdots$} \\
\cmidrule{4-4} \\
& & & \makecell[l]{$\cdots$\textcolor{blue}{{\it he is available for a meeting (or conference call)}}\\\textcolor{blue}{{\it to discuss the GE facility agreement sometime}}\\\textcolor{blue}{{\it  tomorrow - either am or after 300.}} $\cdots$} \\
\cmidrule{4-4} \\
& & & \makecell[l]{Hey Suz\\\textcolor{blue}{{\it Is Sheila still planning on having the GE call}}\\\textcolor{blue}{{\it tomorrow?}}} \\
\cmidrule{2-4} \\
& \multirow{3}{*}[-25pt]{Ph: 000-000-0000} & \multirow{3}{*}[-25pt]{Signature} & \makecell[l]{$\cdots$ Thanks.\\\textcolor{blue}{{\it Rahul}}} \\
\cmidrule{4-4} \\
& & & \makecell[l]{$\cdots$  Larry ********\\\textcolor{blue}{{\it (000) 000-0000}}} \\
\cmidrule{4-4} \\
& & & \makecell[l]{$\cdots$  Director\\\textcolor{blue}{{\it Government Affairs - The Americas}}} \\
 \bottomrule
\end{tabular}
\caption{Nearest Neighbor Analysis on the Enron Dataset. \textcolor{red}{{\bf Red}} denotes the scoped email as predicted by ScopeIt. \textcolor{blue}{{\it Blue}} denotes the actual nearest neighbor in the context. Best viewed in color}
\label{tab:NN}
\vspace{-2mm}
\end{table*}
To generate the set of sentence embeddings, we use the aforementioned publicly available Enron dataset. We randomly sample 10000 of the 500000 emails present in the dataset, and generate sentence embeddings for all the emails, obtaining $\approx 100000$ sentence embeddings. We then generate a typical email that a user might send to their scheduling assistant and use sentences from those emails as query sentences to probe the sentence embedding space. We use Scikit-learn's \cite{scikit-learn} NearestNeighbors method \footnote{\url{https://scikit-learn.org/stable/modules/neighbors.html}} for the NN computation, and retrieve 3 NN sentences.\footnote{Specifically, we retrieve the sentence generating the embedding as well as the email containing the sentence. This is done to provide context, since these sentence embeddings also take context into consideration.} We redact personal information like names, phone numbers in order to preserve the privacy of the users in the Enron dataset.

Table \ref{tab:NN} shows the results of the NN analysis. We use \textcolor{red}{{\bf Red}} to denote the final scoped out email as predicted by ScopeIt, and we use \textcolor{blue}{{\it Blue}} to denote the actual NN of the query sentence. As shown in the table, the queried NNs belong to the same cluster as the query. We see that salutations and signatures get clustered together. We also observe sub-clusters wherein sentences containing date-time availability or phone call intents get mapped to sentences containing similar information. We also observe that contextual information is captured by these contextual embeddings. As shown by the second generated email, syntactically similar query sentences can get mapped to different clusters, based on the context in which they occur.
\end{document}